\documentclass[10pt, conference, letterpaper]{IEEEtran}
\usepackage{xspace}
\usepackage{subfigure}
\usepackage{graphicx}
\usepackage{balance}
\usepackage{comment}
\usepackage{alltt}
\usepackage{multirow}
\usepackage{CJK}
\usepackage{textcomp}
\usepackage{colortbl}
\usepackage{amssymb}
\usepackage{amsmath}
\usepackage{amsfonts}
\usepackage{url}
\usepackage{algorithmic}
\usepackage[ruled,vlined]{algorithm2e}
\usepackage{multicol}
\usepackage{lipsum}
\usepackage{mathrsfs}
\usepackage{booktabs}
\usepackage{makecell}

\graphicspath{{../}{pic/}}
\definecolor{mygray}{gray}{.9}
\definecolor{mypink}{rgb}{.99,.91,.95}
\definecolor{mycyan}{cmyk}{.3,0,0,0}
\newcommand{\ie}{\emph{i.e.}\xspace}
\newcommand{\eg}{\emph{e.g.}\xspace}
\newcommand{\etc}{\emph{etc.}\xspace}
\newcommand{\etal}{\emph{et al.}\xspace}

\newcommand{\Figure}[1]{Fig.~\ref{fig:#1}}
\newcommand{\Equation}[1]{Eq.~\ref{eq:#1}}
\newcommand{\Section}[1]{\S\ref{sec:#1}}
\newcommand{\Table}[1]{Table~\ref{tab:#1}}

\usepackage{booktabs}

\newcommand\system{mmGen\xspace}

\title{One Snapshot is All You Need: A Generalized Method for mmWave Signal Generation}

\author{
 Teng Huang,
 Han Ding*,
 Wenxin Sun,
 Cui Zhao*,
 Ge Wang,
 Fei Wang,
 Kun Zhao,
 Zhi Wang,
 Wei Xi\\
 Xi'an Jiaotong University, China\\
 \{dinghan, zhaocui\}@xjtu.edu.cn
 \vspace{-0.2in}
}

\begin{document}

\maketitle

\begin{abstract}
Wireless sensing systems, particularly those using mmWave technology, offer distinct advantages over traditional vision-based approaches, such as enhanced privacy and effectiveness in poor lighting conditions. These systems, leveraging FMCW signals, have shown success in human-centric applications like localization, gesture recognition, and so on. However, comprehensive mmWave datasets for diverse applications are scarce, often constrained by pre-processed signatures (\eg, point clouds or RA heatmaps) and inconsistent annotation formats. To overcome these limitations, we propose \system, a novel and generalized framework tailored for full-scene mmWave signal generation. By constructing physical signal transmission models, \system synthesizes human-reflected and environment-reflected mmWave signals from the constructed 3D meshes. Additionally, we incorporate methods to account for material properties, antenna gains, and multipath reflections, enhancing the realism of the synthesized signals. We conduct extensive experiments using a prototype system with commercial mmWave devices and Kinect sensors. The results show that the average similarity of Range-Angle and micro-Doppler signatures between the synthesized and real-captured signals across three different environments exceeds 0.91 and 0.89, respectively, demonstrating the effectiveness and practical applicability of \system. 
\end{abstract}

\begin{IEEEkeywords}
mmWave synthesis, wireless sensing
\end{IEEEkeywords}

\section{Introduction}

Wireless sensing systems have recently garnered significant attention due to their distinct advantages over traditional vision-based approaches, such as enhanced privacy protection and resistance to poor lighting conditions, making them promising alternatives. In particular, mmWave-based sensing methods have been attracting increasing interest. The mmWave radar systems, which transmit and receive FMCW signals, offer broader bandwidth and fine-grained sensing capabilities. Leveraging deep learning techniques, these systems have achieved notable success in various human-centric applications \cite{lee2023hupr}\cite{yan2023mmgesture}\cite{lu2020see}\cite{ding2023mi}, including human pose estimation, gesture recognition, localization, and tracking.

Nevertheless, unlike vision-based systems, comprehensive mmWave datasets to support diverse applications remain scarce. Existing datasets face several critical issues: 1) Many datasets \cite{liu2022mtranssee}\cite{mTransSeeDataset} only provide pre-processed mmWave signatures, such as micro-Doppler signatures or point clouds, instead of the original mmWave signals, limiting their applicability to specific sensing tasks. 2) Annotation formats and modalities vary significantly across open-source datasets \cite{mTransSeeDataset}\cite{ge2023large}\cite{wang2024xrf55}. For instance, a dataset intended for motion recognition might include motion labels for each data sample, but this data cannot be readily used for tasks like fine-grained human pose estimation. The process of collecting and labeling mmWave datasets for each sensing task is time-consuming and labor-intensive, which significantly hinders further research and the broader adoption of mmWave sensing technologies.

To address these dilemmas, the literature presents two main types of countermeasures. The initial approach focuses on data augmentation through either signal/image processing \cite{luo2021rfaceid}\cite{chang2024msense} or deep generative models \cite{chi2024rf}\cite{wang2023sudokusens} to enhance the dataset. While this improves certain sensing tasks, it relies heavily on the available data and struggles with zero-shot sensing missions, such as recognizing unseen activities or adapting to new environments.
Recently, attention has shifted towards wireless signal synthesis. This technique holds promise for applications like unseen motion recognition. However, previous methods have limitations: they often focus exclusively on Wi-Fi signals \cite{korany2019xmodal}\cite{RFBoost} or generate only mmWave signatures \cite{bialer2024radsimreal}\cite{Midas}, such as micro-Doppler heatmaps \cite{ahuja2021vid2doppler}. Emerging research has begun to synthesize mmWave signals themselves \cite{xue2023towards}\cite{zhang2022synthesized}, but these efforts typically involve simulating the physical reflections only from the human body and then using deep learning-based refiners to manage environmental attenuation and bridge the gap between synthetic and real signals.
Despite their potential, these methods are not yet practical and universal for widespread use. Each new environment necessitates the training of a new deep learning-based refiner, which in turn requires real-world mmWave data collection and labeling, violating the initial purpose of signal synthesis.

In this paper, we propose a novel software-based framework called \textbf{\system}, a generalized method for \textbf{mm}Wave signal \textbf{Gen}eration tailored for full-scene mmWave signal generation. We build upon the foundational concept of the physical simulator \cite{xue2023towards}\cite{zhang2022synthesized} and extend its capabilities to accommodate arbitrary (or predefined) human motions and physical environments. To realize the above high-level idea, \system faces the following challenges:

1) \textit{Synthesizing Human-Reflected mmWave Signals}. It is complex to determine which parts of the human body reflect mmWave signals and the strength of these reflections upon returning to the radar, especially during diverse human motions. To address this, we represent the human body as a 3D mesh \cite{loper2023smpl} and simulate signal transmission by leveraging the reflection characteristics of mmWave signals on each triangle surface of the human mesh. We utilize the Hidden Point Removal (HPR) algorithm to eliminate occluded meshes and account for the orientation of the remaining surfaces, thereby developing a human reflection signal synthesizer (\Section{human_reflection}). Our approach supports two methods for human mesh construction: one using available video data and the other using textual descriptions of human motions. This flexibility enhances the applicability of \system for various downstream tasks.

2) \textit{Compensating for Environment-Reflected mmWave Signals}. Wireless signals are extremely sensitive to environmental changes, and environments that appear similar to humans can exhibit significant variations in how they profile wireless signals. 
%
%
Therefore, synthesizing environment-reflected signals is crucial to ensure the generated mmWave signals align with practical ones.
Real-world scenarios feature varied layouts and materials of furniture, both of which are critical for mmWave reflections. To address this, we unify the signal synthesizer by constructing a 3D mesh of the environment using a depth camera (\Section{env_reflection}) with only one snapshot and simulating environment-reflected mmWave signals in a manner similar to human reflection signals. Specifically, we consider the reflection coefficient influenced by the material of the furniture and model Tx-Rx antenna gains for refinement. By considering the aforementioned factors, we further enhance the quality of our generated signal.

3) \textit{Tracking Multipath Reflections in Dynamic Scenes}. Multipath effect is also one of the major factors interfering with wireless sensing. Although traditional ray tracing methods can simulate multipath signals, they often entail high time complexity, especially in dynamic scenes. To address this, we leverage the concept of High Reflection Probability Planes (HRPP) to record and track dominant multipath propagations.
This approach strikes a balance between time overhead and quality by simulating multipath reflections among existing furniture and between the furniture and the human body (\Section{multipath_reflection}).

By integrating the reflections mentioned above, \system can generate realistic mmWave signals using a software-based procedure that requires only a single snapshot of the environment. To evaluate the performance of our framework, we built a prototype using commercial off-the-shelf mmWave devices.
%
In summary, the contributions of this paper are:
\begin{itemize}
	\item We model human reflections alongside environmental reflections in a unified manner to generate mmWave radar signals, without relying on deep learning-based methods. Our design enhances the adaptability of various sensing services without the need for extensive data collection for new motions or environments.

	\item We propose a series of techniques to address real-world challenges, such as material properties, antenna gains, and multipath propagations. These techniques make the generated mmWave signal more practical and higher quality.

	\item We develop a prototype system of \system and conduct extensive experiments and evaluations. The results demonstrate the promising performance and practical applicability of our proposed method.
\end{itemize}

\section{Preliminary} \label{sec:pre}

\subsection{Millimeter-wave Radar}

\begin{figure}[t]
    \centering
    \includegraphics[width=.4\textwidth]{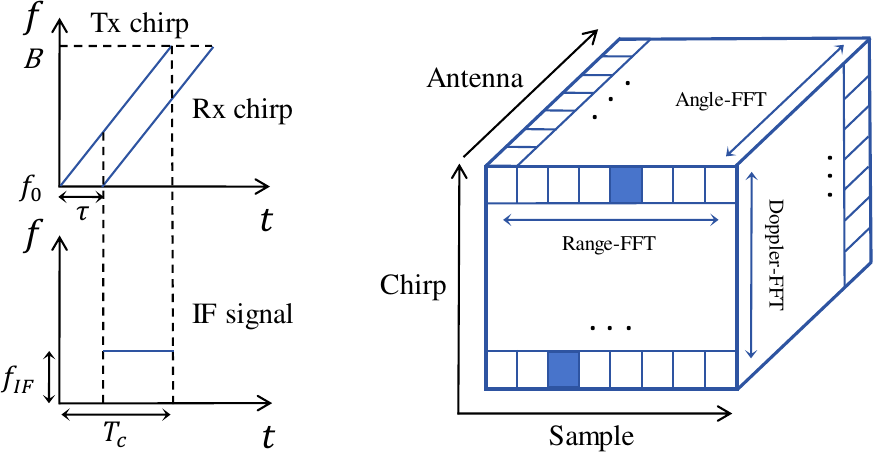}
    \caption{The illustration of FMCW chirp and signal preprocessing.}
    \label{fig:pre_fmcw}
    \vspace{-0.2in}
\end{figure}

The Frequency Modulated Continuous Wave (FMCW) radar is a type of mmWave radar that emits chirp signals with a linearly increasing frequency, as shown in \Figure{pre_fmcw}. Each chirp is characterized by its starting frequency $f_0$, bandwidth $B$, and duration $T_c$. Typically, the radar combines multiple chirps to form a frame, transmitting at a rate of multiple frames per second. Each received chirp has a delay $\tau$ relative to the transmitted one. By mixing the transmitted and received chirps, the radar produces an intermediate frequency (IF) signal with a fixed frequency $f_{IF}$. Analyzing the IF signal allows the radar to measure the range, speed, and angle of objects within its field-of-view (FOV).

\subsection{Sensing Signatures} \label{sec:signatures obtained}

The IF signal is a complex-valued time-domain signal, which undergoes a series of preprocessing steps to obtain signatures for subsequent wireless sensing tasks.
 
\textit{Range-Doppler and Range-Angle heatmaps.} As illustrated in \Figure{pre_fmcw}, a range-FFT is performed along the sample dimension to distinguish objects at different ranges. A doppler-FFT is then performed along the chirp dimension to detect velocity, resulting in the Range-Doppler heatmap. 
Modern mmWave radar systems use multiple transmitting and receiving antennas, forming an antenna array. By performing an angle-FFT on data from these antenna pairs, a Range-Angle heatmap is acquired, showing the relationship between object ranges and their azimuth angles.

\textit{Micro-Doppler heatmap.} The micro-Doppler heatmap describes the relationship between object speeds and fast-time scale within a frame. This spectrogram is derived from the IF signal using Short-Time Fourier Transform (STFT). Unlike the Range-Doppler and Range-Angle heatmaps, the micro-Doppler heatmap represents velocity information over the entire duration of a motion sequence in a single heatmap.

In our evaluation, we analyze the differences between the heatmaps of our generated signals and those of the real captured signals. Furthermore, we train a DNN model using the heatmaps from the generated signals and test it with the real ones to assess their effectiveness in downstream tasks.

\section{Overview}\label{sec:overview}

Our system, \system, is designed to generate practical mmWave signals for sensing scenarios where subjects are involved in various motions. Users can specify these motions or scenarios to generate corresponding datasets for various downstream tasks, such as training deep neural networks for motion recognition, pose estimation, \etc \system supports different input modalities for defining motions: users can provide vision-based datasets depicting motions or describe them textually. Similarly, users can define the sensing scenario, with \system capturing environment details through a snapshot taken by a depth camera.

\begin{figure}[t]
    \centering
    \includegraphics[width=.49\textwidth]{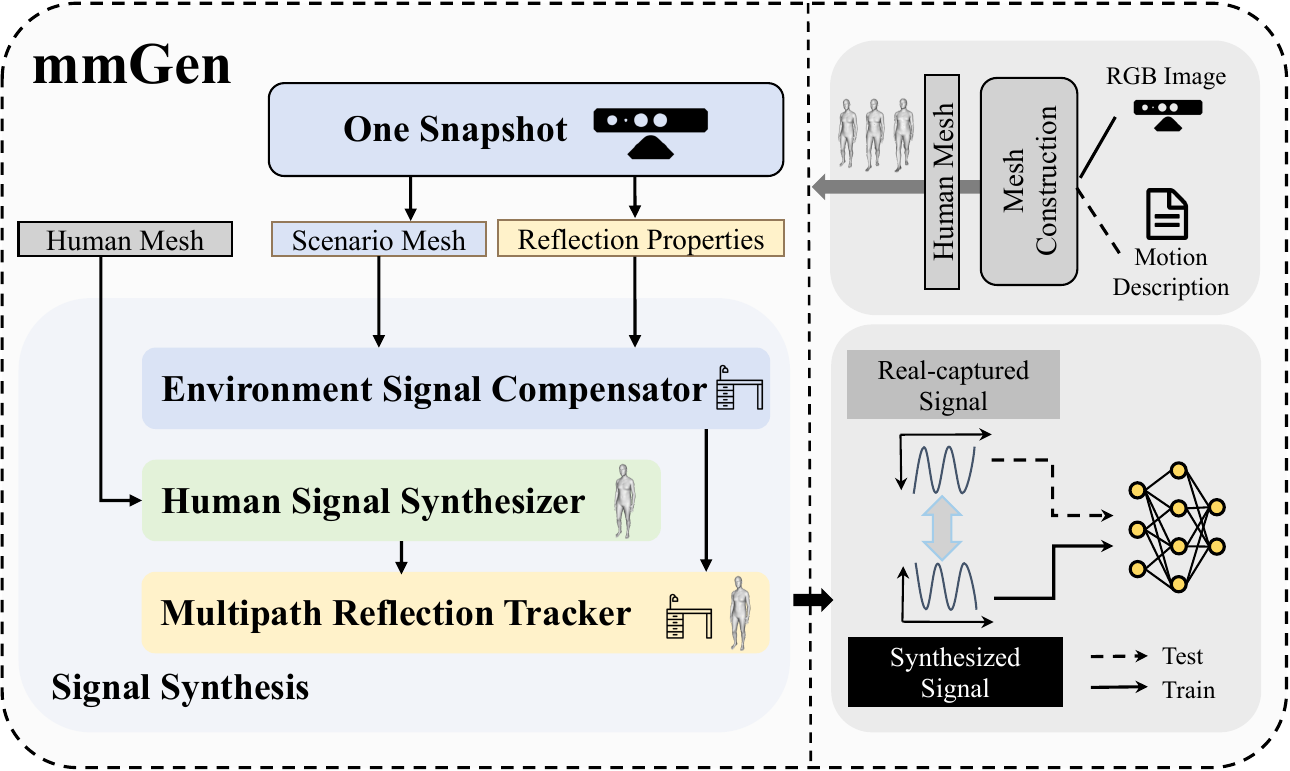}
    \caption{Overview of \system.}
    \label{fig:overview}
        \vspace{-0.15in}
\end{figure}

To achieve these goals, as shown in \Figure{overview}, \system proposes a software-based pipeline that simulates human-reflected (\Section{human_reflection}), environment-reflected (\Section{env_reflection}), and multipath-reflected signals (\Section{multipath_reflection}). 
We construct separate 3D meshes for the human body and the environment, using a unified model to simulate their reflected signals. By decomposing these types of reflections, \system can generate rich mmWave datasets by placing the virtual 3D human mesh at various angles and distances within the environment, significantly reducing dataset collection and labeling costs.
By integrating these reflections, \system provides a generalizable method capable of synthesizing mmWave signals that closely align with real-world collected data. This capability would facilitate various downstream tasks that utilize mmWave signatures for sensing. 

\section{\system Design}\label{sec:design}




In this section, we elaborate on the three main components for mmWave signal generation.

\subsection{Human-Reflected Signal Synthesizer} \label{sec:human_reflection}

We first simulate the human-reflected mmWave signals. In scenarios where people are within the radar's field-of-view (FoV), the radar transmits a continuous frequency-modulated signal, known as a chirp, and receives the reflections from the human body. The radar processes these transmitted and received signals to generate an Intermediate Frequency (IF) signal for sensing. The IF signal reflected from a single point $p_i$ can be represented as:
\begin{equation}
S_i(t) = A_i e^{j2\pi \theta_i}
\end{equation}
where $A_i$ is the attenuated amplitude, and $\theta_i$ is the phase.
To synthesize the signal reflected from the entire human body, we need to identify which points (from the body) will reflect the radar's transmitted signal, and calculate the phase and magnitude of these reflections upon returning to the radar.

Following prior work \cite{xue2023towards}\cite{zhang2022synthesized}, we represent the human body as a 3D mesh which is composed of numerous triangular surfaces. We denote the 3D human mesh leveraging the SMPL model \cite{loper2023smpl}, which profiles the human body with 6890 mesh vertices and 24 joints, allowing us to easily obtain the area and position of each triangular surface. We approximate the centroid of each triangle as the reflection point of the body. 
Since triangles on the reverse side of the human body are blocked and should not contribute to the signal synthesis, we use the Hidden Points Removal (HPR) algorithm to eliminate such triangles. This results in a set of points $\mathbb{P} = \bigcup_{i} \{p_i\}_N$, where $N$ is the number of derived reflection points that are in the line-of-sight to our virtual radar.

With the above operations, we can then compose each IF signal derived from the set $\mathbb{P}$ to obtain the final simulated human-reflected signal:
\begin{equation}
\mathcal{S}_{human}(t) = \sum_{i=1}^N S_i(t) = \sum_{i=1}^N A_i e^{j2\pi \theta_i}
\end{equation}
Then, we need to determine the phase $\theta_i$ and amplitude $A_i$ of each $S_i(t)$.

\subsubsection{Phase calculation}
At time $t$, the phase of the point $i$ is denoted as:
\begin{equation}\label{eq:phase}
\theta_i = f_0 \tau_i  - \frac{B\tau_i^2}{2T_c}  t + \frac{B\tau_i}{T_c}  t
\end{equation}
where $f_0$ is the starting frequency, $B$ is the bandwidth, $T_c$ is the chirp duration, and $\tau_i = \frac{2D}{c}$ is the delay of the received signal relative to the transmitted signal. This delay is determined by calculating the distance between the virtual radar and the centroid of triangle $i$.

\subsubsection{Amplitude calculation}
To calculate the attenuated amplitude, we consider four factors impacting the reflection coefficient of the signal, \ie, the area of the triangle surface $A_{a}$, the orientation of the triangle $A_{o}$, the distance between the reflection point and the virtual radar $D$, and the material of the triangle surface $A_m$.
According to radar communication principles \cite{zhang2022synthesized}\cite{rao2017introduction}, the attenuated amplitude $A_i$ is represented as:
\begin{equation}\label{eq:amplitude}
    A_i =\frac{G_{Tx} G_{Rx} \lambda \sqrt{P} A_{a}^i A_{o}^i A_{m}^i}{(4\pi)^{1.5}D_i^{2}} 
\end{equation}
where $G_{Tx/Rx}$ are the antenna gains, $\lambda$ is the wavelength, and $P$ si the transmission power.
The parameters $G_{Tx/Rx}$ and $P$ are determined by the radar configuration, while $D_i$ and $A_a^i$ can be easily calculated from the body mesh information. Next, we explain how the parameters $A_{o}^i$ and $A_{m}^i$ are calculated.

\begin{figure}[t]
    \centering
    \includegraphics[width=0.5\linewidth]{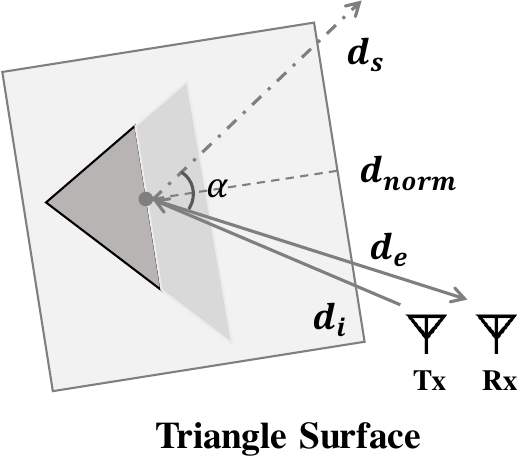}
    \caption{The illustration of how the orientation of a triangle surface affects the reflected signal.}
    \label{fig:des_ao}
    \vspace{-0.15in}
\end{figure}

\textit{Triangle Surface Orientation} ($A_o$): The human body is usually modeled as a quasi-specular reflector \cite{cao2022cross}, where the reflected signals diffuse in different directions according to a Gaussian distribution. As illustrated in \Figure{des_ao}, the strongest reflection of the incident wave $\textbf{d}_{\textbf{i}}$, along with the direction $\textbf{d}_{\textbf{s}}$, is theoretically in the mirror-like reflection direction. If the angle between $\textbf{d}_{\textbf{s}}$ and the reflected wave direction $\textbf{d}_{\textbf{e}}$ is $\alpha$, the reflection coefficient $A_o^i$ is calculated as:
\begin{equation}
A_o^i = e^{-\frac{\alpha^2}{2\eta ^2}}, where\\
\left\{\begin{matrix}
  \textbf{d}_{\textbf{s}}=\frac{\textbf{d}_{\textbf{i}}}{\left \| \textbf{d}_{\textbf{i}}\right \| }  - 2\frac{\textbf{d}_{\textbf{i}}\cdot \textbf{d}_{\textbf{norm}}}{\left \| \textbf{d}_{\textbf{i}}\right \|} \textbf{d}_{\textbf{norm}}\\
\alpha =\arccos \frac{\textbf{d}_{\textbf{e}} \cdot \textbf{d}_{\textbf{s}}}{\left \| \textbf{d}_{\textbf{e}} \right \| } 
\end{matrix}\right.
\end{equation}
where $\eta$ is an empirical value, and $\textbf{d}_{\textbf{norm}}$ is the unit vector in the direction of the normal. 
Thus, when the locations of the virtual radar and the human body are fixed, $A_o^i$ is determined by the orientation of each triangle surface.

\textit{Triangle Surface Material} ($A_m$): The material of the triangle surface affects the signal amplitude when mmWave signals propagate to the interface of two different media. The relationship between the reflected and incident waves is modeled using the Fresnel reflection principle \cite{Paschottafresnel_reflections}. The Fresnel reflection coefficients are represented by the vertical polarization coefficient $\Gamma_{v}$ and the horizontal polarization coefficient $\Gamma_{h}$, influenced by the incidence angle of the radar signal and the complex permittivity of the surface:
\begin{equation}
    \Gamma_{v}^i=\frac{\varepsilon_i \sin\beta_{i} - \sqrt{\varepsilon_i-(\cos\beta_{i} )^{2}}  }{\varepsilon_i \sin\beta_{i} + \sqrt{\varepsilon_i-(\cos\beta_{i} )^{2}}}  
\end{equation}
\begin{equation}
    \Gamma_{h}^i=\frac{\sin\beta_{i} - \sqrt{\varepsilon_i-(\cos\beta_{i} )^{2}}  }{\sin\beta_{i} + \sqrt{\varepsilon_i-(\cos\beta_{i} )^{2}}}
\end{equation}
where $\beta_{i}$ is the incidence angle, and $\varepsilon_i$ is the complex permittivity of the surface $i$, given by $\varepsilon_i = \varepsilon_i^{'}-j60\lambda\sigma_i$, with the relative permittivity $\varepsilon_i^{'}$, wavelength $\lambda$ and the conductivity of the medium $\sigma_i$. 
Finally, we calculate the reflection coefficient related to the material as:$A_{m}^i = \Gamma_{v}^i\hat{v} +\Gamma_{h}^i\hat{h} $, where $\hat{v}$ and $\hat{h}$ represent unit vectors of the electric field components perpendicular and parallel to the plane of incidence, respectively. By incorporating these material-related impact factors, we can simulate the reflected signals more accurately.

Due to significant variations in electromagnetic properties among different individuals and even among different parts of the same human body, we set $\varepsilon^{'}$ as 10 and $\sigma$ as 1e-10 experimentally for human surfaces is adequate. Interestingly, in the simulation of environment reflection signals in \Section{env_reflection}, the material differences of furniture introduced by $A_m$ greatly enhance the quality of the synthesized signal.

\subsection{Environment-Reflected Signal Compensator}\label{sec:env_reflection}

\begin{figure}[t]
    \centering
    \includegraphics[width=0.45\textwidth]{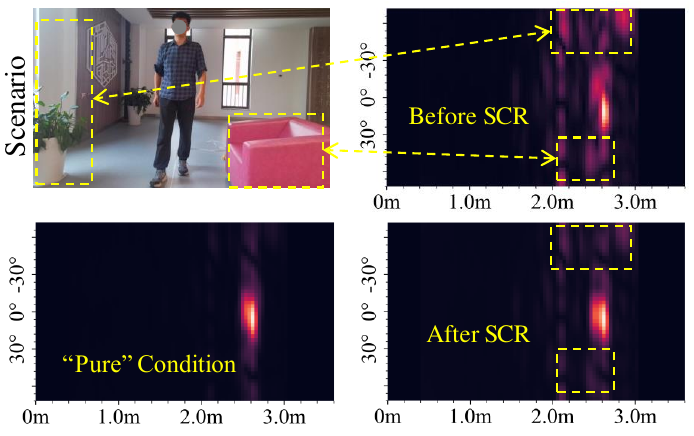}
    \caption{RA heatmaps before/after performing the static clutter removal.}
    \label{fig:des_scr}
    \vspace{-0.15in}
\end{figure}

RF signals are highly sensitive to environmental changes, making it essential to adapt a sensing model for specific environments to ensure real-world applicability. Our \system system simulates environment-reflected signals, aligning the synthesized signals with real-world scenarios, thus providing valuable datasets for enhancing the sensing model. While some may argue that environmental factors are constant since furniture typically does not move, static clutter removal (SCR) algorithms can reduce interference and make the sensing model less dependent on signals with environment reflections. However, our experiments and observations (illustrated in \Figure{des_scr}) demonstrate that SCR only partially mitigates the impact when there is human motion within the scenario. Therefore, simulating the environmental reflections of a real scene and incorporating these signals can more effectively support the specific sensing model, as evidenced in the evaluation (\Section{ablation}).

\subsubsection{Scene Mesh Construction}
To simulate environment-reflected signals, we construct a 3D mesh of the scene, similar to the human body representation. This method allows us to reuse the signal synthesizer designed for human reflections. We use a depth camera to capture a snapshot of the environment, obtaining the point clouds. The normal estimation and mesh construction methods from the open-source library Open3D \cite{zhou2018open3d} are then used to build the mesh from the point cloud. The normal estimation method calculates the principal axis of adjacent points using covariance analysis, while the mesh construction method, based on the ball pivoting algorithm, dynamically adjusts the ball radius to form triangle planes in the mesh. After constructing the environment mesh, we record the normal vector and area of each triangle surface, which are related to the signal amplitude during signal generation. We then calculate the signal phase, and adjust the amplitude parameters, which is depicted in the following.

\subsubsection{Furniture Material}

Furniture in the target scene varies in material, affecting the reflected signals. To differentiate and set $A_m$ accordingly, we use a pre-trained VoteNet model \cite{qi2019deep} to detect 3D objects in the scene. VoteNet is an end-to-end detection model suitable for indoor scenes, providing category labels, confidence scores, and spatial bounding boxes for each object. By matching detected objects with their materials using online databases \cite{matweb}, we determine the electromagnetic reflection properties ($\varepsilon^{'}$ and $\sigma$) of each object\footnote{To demonstrate the effectiveness of our approach, we incorporate seven materials in the implementation, including plywood panels, polyurethane, paperboard, ceramic, glass, concrete, and leather. These materials can be expanded and adjusted as needed in real-world applications.}. The derived $A_m^i$ is then used to incorporate the variation of material factors when calculating the signal amplitude.
%

\subsubsection{Antenna Gains}
The gain of the radar antenna varies with the angle of the reflecting object relative to the radar, significantly impacting the generated signals. Previous works \cite{xue2023towards}\cite{zhang2022synthesized} often overlooked this factor, as they only simulate human reflections and assume the human locates directly in front of the radar. However, environmental objects are often located at various azimuth angles. If the variation in antenna gain is not considered, the generated signals will deviate from reality (as demonstrated in \Table{ablation}). According to \cite{IWR6843L}, the signal gain of any Tx-Rx antenna pair and the azimuth angle of the reflecting object approximately follow a Gaussian distribution. Referring to the mmWave radar user's guide \cite{IWR6843L}, we determine the standard deviation of this Gaussian distribution by curve fitting, obtaining the gain function $G_{a}(\varphi_a)$ for azimuth angles and $G_{e}(\varphi_e)$ for elevation angles.

We incorporate the furniture material factor and gain functions into the amplitude calculation for environment-reflected signals of triangle surface $i$: 
\begin{equation}
    A_i^{env} =\frac{G_{Tx} G_{Rx} G_{a}(\varphi_a^i) G_{e}(\varphi_e^i) \lambda \sqrt{P} A_a^i A_o^i A_m^i}{(4\pi)^{1.5}D_i^{2}} 
\end{equation}
In summary, we reuse the human body signal synthesizer to generate signals reflected from environmental objects $S_{env}(t)$. The only difference is replacing $A_i$ with $A_i^{env}$. 

\subsection{Multipath Reflection Tracker}\label{sec:multipath_reflection}

In indoor environments, the multipath effect is nonnegligible. Inspired by \cite{chen2023differentiable}, which uses Monte Carlo Sampling for ray tracing \cite{dudek2010millimeter} in a 3D scene to calculate reflection signals, we analyze radar signal propagation paths and generate the corresponding multipath reflections.
Due to the high center frequency and strong directivity of commercial mmWave radars, the propagation attenuation in space is rapid. Our simulation results for human- and environment-reflected signals indicate significant attenuation after a single reflection. Thus, to simplify the simulation of multipath reflections, we explore only situations where the signal returns to the radar after two reflections. We use High Reflection Probability Planes (HRPP) to record these propagation paths for subsequent signal generation.

\begin{figure}[t]
    \centering
    \includegraphics[width=1\linewidth]{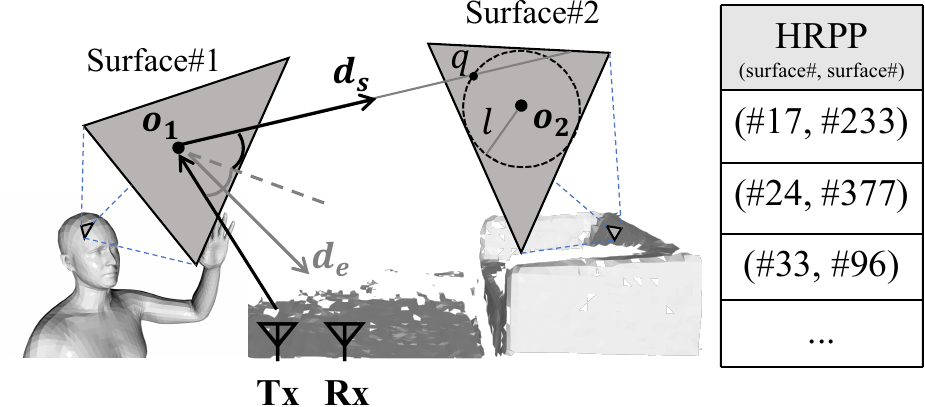}
    \caption{Multipath reflection tracing and HRPP table construction.}
    \label{fig:des_hrrp}
    \vspace{-0.15in}
\end{figure}

As shown in \Figure{des_hrrp}, the radar signal bounces off from triangle surface\#1, producing a reflection towards the direction $\textbf{d}_{\textbf{e}}$ to the Rx antenna. We denote the direction of the strongest reflection as $\textbf{d}_{\textbf{s}}$, and trace the second reflection emitted along this path. 
If triangle surface\#2 contains an inscribed sphere $l$ and the ray $\textbf{d}_{\textbf{s}}$ intersects this sphere at point $q$, a second reflection along $\textbf{d}_{\textbf{s}}$ is confirmed. Point $q$ is acquired by solving the following equation for $k$:
\begin{equation}
    (\textbf{o}_{\textbf{1}}+k\textbf{d}_{\textbf{s}} -\textbf{o}_{\textbf{2}})^{2}-l^{2}=0
\end{equation}
Once determined, this reflection surface pair $i$ is added into our HRPP table.
The propagation path of this multipath reflection is `Tx$\rightarrow$Surface\#1$\rightarrow$Surface\#2$\rightarrow$Rx'.
To simulate the multipath signal $S_{multipath}(t)$, we consider the cumulative gain of reflections along the path.

In detail, the phase is calculated following \Equation{phase}, where the distance parameter is replaced with the total path length. 
For amplitude calculation, we revise the attenuated amplitude as:
\begin{equation}
    A_{i}^{multipath} = \frac{G_{Tx} G_{Rx} G^{{pair}_i} \lambda \sqrt{P} A_a^{{pair}_i} A_o^{{pair}_i} A_m^{{pair}_i} }{(4\pi)^{1.5}D_{{pair}_i}^{2}}
\end{equation}
where $G^{{pair}_i}$, $A_a^{{pair}_i}$, $A_o^{{pair}_i}$, and $A_m^{{pair}_i}$ are the products of the corresponding parameters of the two surfaces respectively.
During the generation phase, we maintain an HRPP table where each element corresponds to a pair of surfaces. This table is updated whenever the reflection situation changes (\ie, when the human moves). This approach allows us to simulate multipath signals in dynamic scenarios, balancing computational complexity and signal fidelity.

\textbf{Summary}: we combine the human-reflected, environment-reflected, and multipath signals to acquire the final synthesized mmWave signal:
\begin{equation}\label{eq:Ssyn}
S_{syn} = S_{human} + S_{env} + S_{multipath}
\end{equation}
It is important to note that \Equation{Ssyn} represents the simulation result from a single pair of Tx and Rx antennas. Depending on the actual radar configuration, we will simulate the signals for multiple radar transceiver pairs accordingly.

\section{Implementation}\label{sec:implementation}

\subsection{Experimental Setups}

\textbf{mmWave Testbed:} We use the commercial TI IWR6843isk mmWave radar jointly with the DCA1000EVM to collect real-world signals. These signals are transmitted to a laptop via the UDP protocol. The radar features 3 Tx antennas and 4 Rx antennas, and operates with a bandwidth of 4$GHz$. The radar's starting frequency $f_{c}$ is 60$GHz$, with a chirp ramp time $T_{c}$ of 28$\mu$s and an idle time of 7$\mu$s. The frame rate is set to 15$fps$. Each frame consists of 255 chirps, with 256 points sampled per chirp. Under these configurations, the maximum sensing range is 10.49$m$, with a maximum speed of 11.9$m/s$, a range resolution of 0.041$m$, and a speed resolution of 0.093$m/s$.

\textbf{Kinect:} The Kinect device, equipped with a depth camera, an RGB camera, and motion sensors, is configured to operate at a frame rate of 15$fps$ with a resolution of 720$p$ in wide field-of-view (WFoV) binned mode. We use this device to capture the depth image of the scene, convert it to point cloud, and generate the 3D mesh of the scene. Specifically, when generating the mesh, we set the normal estimation search radius to 10$cm$ and the maximum number of neighbors to 50. The ball pivoting algorithm's ball radius is set to 1.5 times the average vertex spacing, resulting in an environment mesh with an average surface spacing of 3.9$cm$, which closely matches the mmWave range resolution. 
Additionally, the Kinect captures synchronized RGB images while collecting mmWave signals. 

\begin{figure}[t]
    \centering
    \includegraphics[width=.8\linewidth]{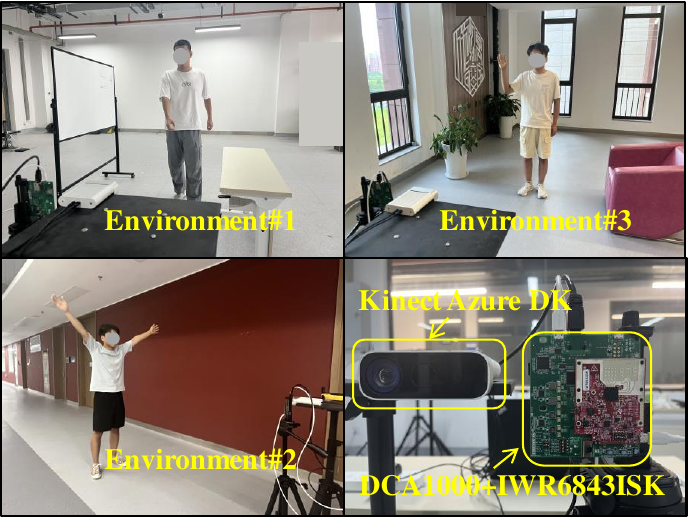}
    \caption{Experimental scenarios and testbeds.}
    \label{fig:imp_env}
    \vspace{-0.1in}
\end{figure}

\textbf{Scenes}: Both the Kinect and the mmWave radar are placed at a height of about 1.1$m$ from the ground, with objects and people positioned within about 3$m$ of the radar. We deploy our testbed in three environments (\Figure{imp_env}) to conduct the experiments. In Env\#1, we use common indoor furniture such as whiteboards and desks as reflective surfaces. Env\#2 is a corridor with  walls serving as reflectors. Env\#3 include furniture such as sofas, windows, and potted plants. Additionally, the floors in all three scenarios contribute to the reflective signals. The variety in shapes and materials of these indoor items is sufficient to validate the effectiveness of our proposed method.

\subsection{Datasets}
We invited six volunteers (four males and two females) to perform a set of six motions in front of the mmWave radar and Kinect, capturing both mmWave signals and RGB images. The six selected human motions are: 1) Walking, 2) Clapping, 3) Waving left hand, 4) Jumping jacks, 5) Picking up a stone, and 6) Turning around. These motions are sourced from the HumanML3D dataset \cite{guo2022generating}, which is used to pre-train the diffusion-based MDM model \cite{tevet2022humanmotiondiffusionmodel} for 3D human mesh construction. The chosen motions cover upper body movements, slow full-body movements, vigorous full-body movements, and position-changing actions, providing certain diversity. We created three datasets for evaluation:

$\mathbb{D}1$: This dataset contains real-captured mmWave signals, used exclusively in the testing phase to evaluate our system. 

$\mathbb{D}2$: We generated 3D human meshes from the RGB images using the Hand4Whole model \cite{Moon_2022_CVPRW_Hand4Whole} and constructed the environment mesh from the Kinect-collected point cloud. These meshes were then processed through \system to generate the synthesized mmWave signals, forming dataset $\mathbb{D}2$.

$\mathbb{D}3$: Textual descriptions of the six motions were input into the diffusion-based MDM model \cite{tevet2022humanmotiondiffusionmodel} to generate 3D human meshes. These meshes, combined with the Kinect-constructed environment mesh, were processed through \system to create dataset $\mathbb{D}3$. Specifically, we generated 16.2$k$ frames of human mesh data, with each sample consisting of 45 frames. Different shape parameters were used for various samples to enhance dataset diversity. Prior to inputting into \system, we interpolated between adjacent frames to match the number of mmWave chirps. Ultimately, dataset $\mathbb{D}3$ contains 48.6$k$ frames with 12.3 million chirps.

\section{Evaluation}
In this section, we evaluate \system from two perspectives: signal similarity and effectiveness in the downstream task.

\subsection{Metrics}

\textbf{MS-SSIM:} Multi-Scale Structural Similarity (MS-SSIM) is an evaluation metric that ranges from 0 to 1, measuring the structural similarity of two samples at multiple scales. 
%
%
This metric assesses whether our method accurately reproduces real-captured mmWave signals.

\textbf{Confusion Matrix:} For the activity recognition task, we train a model using dataset $\mathbb{D}3$ and test it on the real-captured dataset $\mathbb{D}1$. The confusion matrix evaluates the recognition performance, illustrating how well the model distinguishes between different activities.


\subsection{Overall Generation Quality}

\subsubsection{Signature Similarity}

We first evaluate the similarity between our generated signals ($\mathbb{D}2$) and the corresponding real-captured signals ($\mathbb{D}1$) from both qualitative and quantitative perspectives.

\begin{figure*}[htbp]
\centering
\includegraphics[scale=.65]{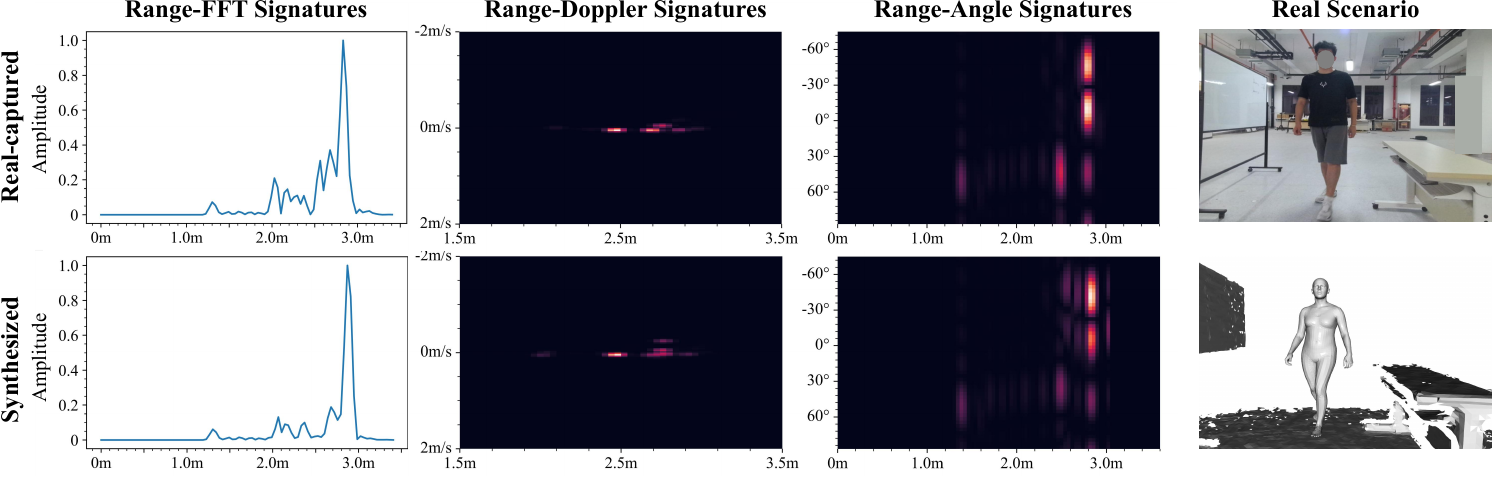}
\caption{Qualitative evaluation on three common signal signatures between the real-captured signal and the generated signal.}
\label{fig:eval_qualitative}
\vspace{-0.15in}
\end{figure*}

\textbf{Qualitative Results}. \Figure{eval_qualitative} illustrates the qualitative result, where the rightmost part displays the real scenario and mesh of a volunteer performing a `walk' motion during the experiment. Based on our mmWave radar configuration, we capture reflected signals from the human body, board, table, and a small portion of the ground in this scenario. The Kinect device synchronously captures RGB images and timestamps. We randomly select one frame, use the corresponding RGB image to acquire the human mesh, and the depth point cloud to acquire the environment mesh. We then use \system's pipeline to synthesize the mmWave signal. After preprocessing both the generated and real-captured signals, we display their signatures, including Range-FFT, Range-Doppler (RD), and Range-Angle (RA) signatures. All signatures are normalized in amplitude and shown on the the left side of \Figure{eval_qualitative}. 

The results indicate that all types of signatures for the generated signals are very similar to the real-captured signals, demonstrating the practical signal synthesis capability of our proposed \system. Notably, the RA signature clearly shows that the signals generated by \system can reproduce the reflections from objects (\eg, board and table) in the real scenario, with high signal strengths appearing outside the human positions. This accuracy is attributed to the compensation for environmental and multipath reflections.

\begin{figure}
\centering
\subfigure[RA signature]
{\includegraphics[width=0.24\textwidth]{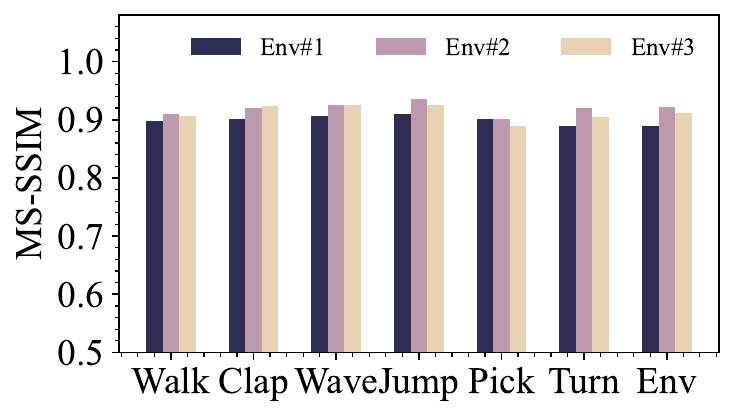}}
\subfigure[MD signature]
{\includegraphics[width=0.24\textwidth]{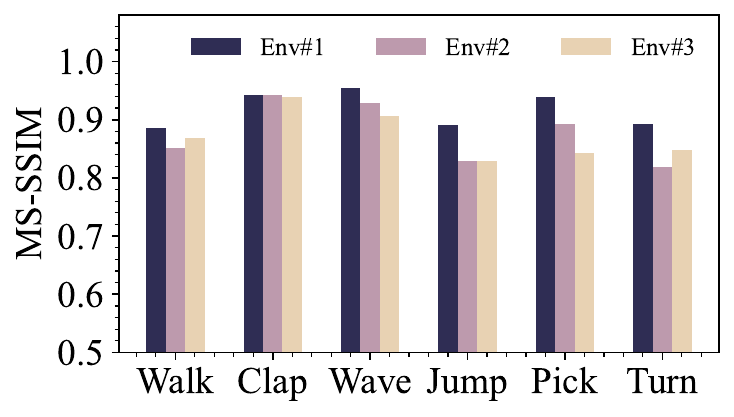}}
\caption{The similarity of RA and MD signatures between the real and the generated signals.}
\label{fig:eval_overall_ssim}
\vspace{-0.15in}
\end{figure}

\textbf{Quantitative Evaluation.} In this part, we quantitatively evaluate the quality of the RA and micro-Doppler (MD) signatures of the generated signals using the MS-SSIM metric. 

\textit{RA signatures}: As shown in \Figure{eval_overall_ssim}(a), the mean MS-SSIM across all motions in the three scenarios is 0.91. Regardless of the range or intensity of the motions, \system effectively reproduces the signals. Signal quality is slightly higher in the second scenario (Env\#2), due to the simpler environment with only wall and floor reflections. Specifically, the last group of bars displays the results when only environmental reflections are present, with the MS-SSIM still exceeding 0.90.

\textit{MD signatures}: \Figure{eval_overall_ssim}(b) shows that despite there are fluctuations in signal quality across different motions and environments, the mean MS-SSIM is 0.89, demonstrating \system's ability to capture dynamic information. For motions such as `Jumping jack',  `Pick up a stone', and `Turn around', which involve significant movements of certain body parts, our method still effectively captures the overall dynamic characteristics of the signals. This success is due to \system's dynamic tracking of reflections at the chirp level, ensuring fine-grained dynamic information.

\begin{small}
\begin{table}[tbp]
\centering
  \caption{MAE and SD on Micro-Doppler Signature.}
    \vspace{-0.1in}
  \label{tab:comparison}
  \begin{tabular}{cccccllll}
    \toprule
    \multicolumn{2}{c}{}     &  MAE      &  SD      \\
   \hline
    \multirow{3}{*}{\textbf{\system (Ours)}} & Env\#1  &  0.05  &  0.01     \\
    & Env\#2               &  0.05       & 0.01          \\
    &Env\#3              &  0.06       & 0.01          \\
  \hline
   \multicolumn{2}{c}{Midas \cite{Midas}}        & 0.06  & 0.02 \\
 \hline
  \multicolumn{2}{c}{Vid2Doppler \cite{ahuja2021vid2doppler}}       & 0.09 & 0.03  \\
  \bottomrule
\end{tabular}
\vspace{-0.15in}
\end{table}
\end{small}

\subsubsection{Comparisons}
Next, we compare our work with two prior methods, \ie, Midas \cite{Midas} and Vid2Doppler \cite{ahuja2021vid2doppler}, which also generate MD signatures. They use video data as input and leverage deep learning to generate the corresponding MD signatures. We calculate the Mean Absolute Error (MAE) and Standard Deviation (SD) for the generated and real signals in three scenarios. 
Due to different radar settings, generated signatures vary in size. For a fair comparison, we extract a portion of our signatures matching the size (32 radial velocity bins $\times$ 75 frames) used in these methods. 
%
As the results listed in \Table{comparison}, our metrics outperform these two approaches across all scenarios. Unlike these methods, which rely on deep neural networks during signal generation, \system maintains generalization and adapts to different scenarios.

\subsubsection{Ablation Study}\label{sec:ablation}

Then, we conduct an ablation study to investigate the effectiveness of each module in our signal generation process. The similarity of RA and MD is evaluated, and the results are shown in \Table{ablation}.

`w/o B\&C' represents without considering any environmental reflections (\Section{env_reflection} and \Section{multipath_reflection}). We can see that with this setting, the average RA similarity drops from 0.90 to 0.85. This significant decrease is reasonable because environmental and multipath signals from objects are crucial components of the RA signature, which confirms the necessity of simulating these reflections.

`w/o $A_{m}$' represents we only include the environment mesh in the reflection path without considering their material factors. We see that the RA similarity decreases to varying extents across the three scenes, especially in Env\#3, which contains materials like ceramics and leather. This confirms the necessity and effectiveness of material factor estimation, as ignoring material properties would result in high amplitudes from small reflective surfaces, reducing signal quality.

`w/o $G_{a/e}$' represents excluding antenna gain effects when calculating environmental reflections. The results show that RA similarity decreases significantly. According to the radar handbook \cite{IWR6843L}, reflections from objects at azimuth angles beyond ±60 degrees experience significant attenuation compared to objects directly facing the radar. Env\#1, with more reflective surfaces near ±60 degrees, is most affected by the absence of $G_{a/e}$, proving the significance of our $G_{a/e}$ modeling.

We also observe that Env\#2, with simpler reflection paths, is least affected by the absence of different modules compared to the other scenarios, which implicitly indicates the importance of modeling environmental reflections for complex scenes. 

In comparison, MD similarity is less impacted. This is because static environmental reflections are eliminated during preprocessing (by the static clutter removal algorithm). However, similarity still decreases when different design components are excluded. The reason is that \system tracks multipath reflections that cannot be eliminated in dynamic scenarios.

\begin{small}
\begin{table}[tbp]
\centering
  \caption{Ablation Study.}
    \vspace{-0.1in}
  \label{tab:ablation}
  \setlength{\tabcolsep}{2.45pt}
  \begin{tabular}{c|cccc|ccccllll}
    \toprule
    \multicolumn{1}{c|}{} & \multicolumn{4}{c|}{RA} & \multicolumn{4}{c}{MD}\\
     & Env\#1 &Env\#2&Env\#3&Avg  &  Env\#1 &Env\#2&Env\#3&Avg  \\
     \Xcline{1-1}{0.4pt}
   \hline
    \textbf{\system (Ours)} & 0.90& 0.92&  0.91&  0.91 &0.92&  0.88&  0.87&  0.89\\
    w/o B\&C  & 0.85& 0.87&  0.84&  0.85& 0.89&  0.87&  0.87 & 0.88  &    \\
    w/o $A_m$   &  0.85& 0.88&  0.81&  0.85& 0.89&  0.88& 0.85&  0.87\\
    w/o $G_{a/e}$  & 0.82& 0.85& 0.84&  0.84& 0.88 & 0.86&  0.85&  0.87  \\
  \bottomrule
\end{tabular}
\vspace{-0.15in}
\end{table}
\end{small}


\subsection{Micro-benchmarks}

\subsubsection{Impact of Distance}

We test \system's signal quality when the human subject positioned at 1$m$, 2$m$, and 3$m$ from the radar. As shown in \Figure{eval_benchmark}(a), our method generates high-quality signals at all distances. However, performance fluctuations slightly increase with distance. This can be attributed to two factors: 1) Real signals attenuate over longer distances, causing amplitude differences between real and generated signals, and 2) Greater distances introduce more variations and fluctuations in real-captured signals between frames, increasing metric variability.

\subsubsection{Impact of Angle}

\begin{figure}[tbp]
\centering
\subfigure[Distance]
{\includegraphics[width=0.24\textwidth]{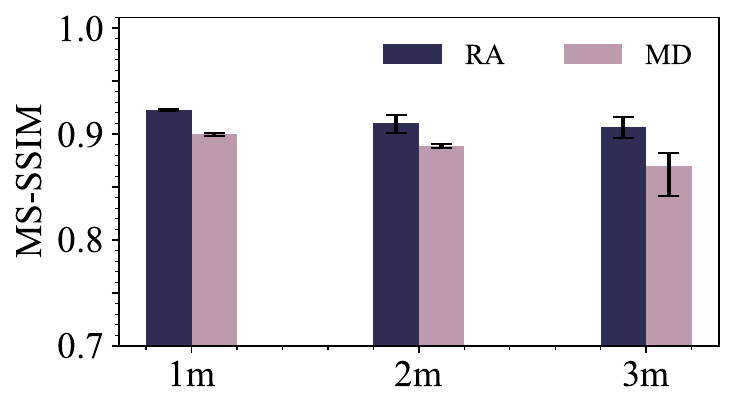}}
\subfigure[Angle]
{\includegraphics[width=0.24\textwidth]{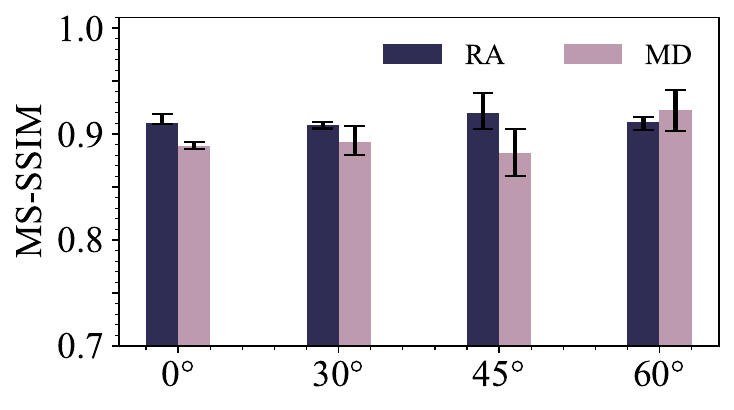}}
\vspace{-0.05in}
\caption{The performance under different experimental conditions.}
\label{fig:eval_benchmark}
\vspace{-0.15in}
\end{figure}

We also test \system's signal quality at various human body rotation angles: 0 degrees, 30 degrees, 45 degrees, and 60 degrees. As shown in the \Figure{eval_benchmark}(b), our method consistently produces synthetic signals similar to real-captured signals at all angles.
While it is widely accepted that mmWave signals perform poorly across angles in downstream sensing tasks, this limitation is task-specific rather than an issue of signal modeling. \system models the signal from a physical perspective, accurately tracking propagation paths regardless of the orientation of reflective subjects. As a result, our synthesized signals are slightly impacted by changes in angle.

\subsubsection{Time Efficiency}
Here, we investigate the time efficiency of \system on a desktop with an Intel Core i5 8500 CPU.
With proper program optimizations, \system can generate a single frame of mmWave signal (comprising 255 chirps) within 6$s$. This process includes 3.27$s$ for modeling human reflections, 0.12$s$ for simulating environment reflections, and 2.12$s$ for simulating multipath reflections, all for 12 virtual antenna pairs. We believe this simulation process is cost-effective, especially given the superior generalization capability of our system, which significantly reduces the labor required for real data collection.

\subsection{Case Study}

In this section, we demonstrate how \system can enhance a common downstream mmWave sensing task, \ie, activity recognition.

\textbf{Setup}. We use the generated mmWave signals from dataset $\mathbb{D}3$ as training data. This dataset includes all 6 motions, each performed 60 times in various scenarios, with each activity lasting 3 seconds (45 frames). 
The real-captured mmWave data from dataset $\mathbb{D}1$ is used for testing. 
This setup aims to demonstrate that our generated signals are realistic enough for common activity recognition tasks, effectively bridging the gap between synthetic and real data. Notably, there is no overlap between the human subjects in the training set $\mathbb{D}3$ (where the human mesh is generated virtually using motion descriptions) and the test set $\mathbb{D}1$, making this a natural cross-domain test and implicitly increasing the motion recognition difficulty.

\textbf{DNN structure}. For each sample, we obtain its multi-frame RA signatures and fast-time dimension MD signatures as inputs to a designed DNN network for activity recognition. 
The DNN is a multi-layer neural network consisting of two CNN-based encoders, a bidirectional LSTM module, and an MLP module. The encoder includes five CNN blocks with residual connections to enhance feature extraction. This is followed by a two-layer bidirectional LSTM to further capture the temporal information of activities. Finally, an MLP module classifies the activities. The network uses cross-entropy loss, with the Adam optimizer and a learning rate of 2e-5.

\begin{figure*}[t]
\centering
\subfigure[Overall performance]
{\includegraphics[width=0.24\textwidth]{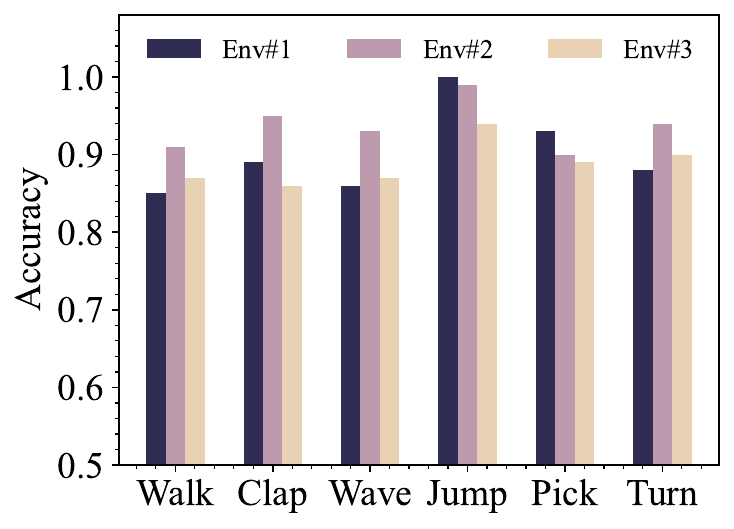}}
\subfigure[Env\#1]
{\includegraphics[width=0.22\textwidth]{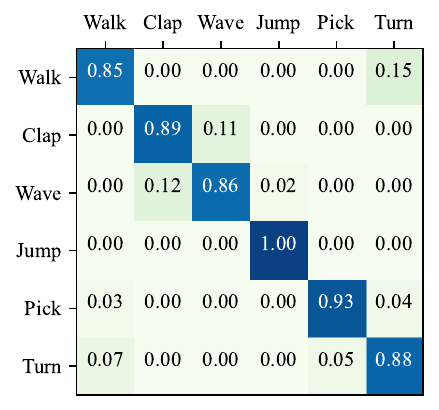}}
\subfigure[Env\#2]
{\includegraphics[width=0.22\textwidth]{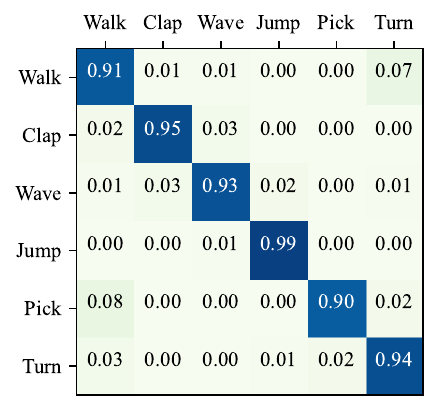}}
\subfigure[Env\#3]
{\includegraphics[width=0.22\textwidth]{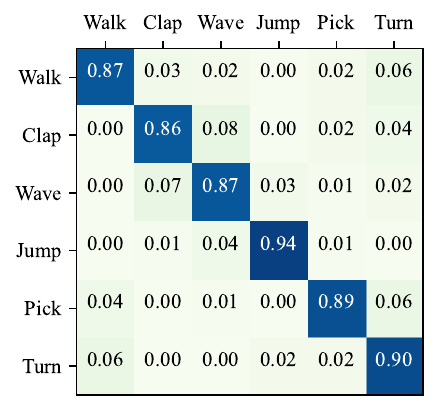}}
\caption{The performance of activity recognition.}
\label{fig:exp_activity_accu}
\vspace{-0.15in}
\end{figure*}

\textbf{Results}. As shown in \Figure{exp_activity_accu}, we train and test the model in three different scenarios, presenting overall accuracy and the confusion matrices for each environment.
This experiment demonstrates that the signals generated by \system carry sufficient information for human activity profiling. Our classification accuracy reaches 91\%, which is impressive considering that we did not use any real-captured mmWave samples during training. The consistent classification accuracy across the three environments also indicates that our generation method generalizes well to various environments, aligning with our expectations.

We observe that `Jumping jack' has a high recognition accuracy in all environments. The unique movement pattern of this activity results in distinctive signatures, making it easier to recognize compared to other activities. Conversely, `Walk' and `Turn around' exhibit some degree of confusion in all three environments. By diving into the training data in $\mathbb{D}3$, we found that many `Turn around' samples initially involved a short `walk' period, causing a shift in signature distribution. This shift, arising from the mesh generation process (using the diffusion-based MDM model), confuses the neural network during learning, leading to reduced recognition accuracy. We believe this issue can be mitigated by utilizing a more advanced mesh generation model, which is beyond the scope of this work.

\section{Related Work} \label{sec:related}

Wireless signal synthesis has emerged as an effective method to address the scarcity of datasets in wireless sensing tasks, garnering significant research attention recently. Initial efforts in the field have primarily focused on WiFi signal augmentation \cite{korany2019xmodal}\cite{RFBoost}\cite{Cai_Korany_Karanam_Mostofi_2020}. For instance, RF-Boost \cite{RFBoost} introduces a plug-and-play framework that augments WiFi time-frequency spectrograms using time-, frequency-, and space diversity before applying deep neural network (DNN) learning. XModal-ID \cite{korany2019xmodal} simulates WiFi signals by physically modeling reflections on the human body's surface. Additionally, Cai \etal \cite{Cai_Korany_Karanam_Mostofi_2020} simulate WiFi signals directly from video streams for gym activity classification. 
%
%
However, these approaches are specific to WiFi signal augmentation and cannot be directly applied to FMCW mmWave signals. Moreover, the generated data often lack the necessary diversity and generalization capabilities, limiting their broader applicability.

In recent years, mmWave-based sensing systems have shown promising performance, highlighting the need for building comprehensive datasets to promote community progress. Regarding mmWave signal generation, the pioneering Vid2Doppler \cite{ahuja2021vid2doppler} synthesizes micro-Doppler heatmaps using publicly available unstructured videos. Building on this, Midas \cite{Midas} incorporates multi-person scenarios and multipath effects. However, these methods are task-specific, generating only mmWave signatures such as micro-Doppler heatmaps or point clouds.

Recent research has begun synthesizing original mmWave signals. SynMotion \cite{zhang2022synthesized} models physical reflections from the human body, calculating reflected signals based on radar transmission paths to synthesize raw signals. mmGPE \cite{xue2023towards} models reflections from different parts of the human body and uses a GAN model \cite{goodfellow2014generativeadversarialnetworks} to refine signal signatures. While these methods demonstrate the feasibility of generating raw radar signals from a physical modeling perspective, they only consider reflections from human body, neglecting significant environmental factors. Consequently, they rely on DNN models to bridge the gap between real and synthetic data.

More recent studies \cite{chi2024rf}\cite{chen2023differentiable}\cite{tilemask} have started to generate radar signals reflected from objects, analyzing the influencing factors of object surface reflections. However, these approaches have limitations. For instance, TileMask \cite{tilemask} only considers metal reflections, while DiffSBR \cite{chen2023differentiable} and RF-Diffusion \cite{chi2024rf} rely on deep learning methods to estimate signal properties reflected from various objects in the specific scenario. This makes the synthesis process a black-box and difficult to generalize cost-effectively to other scenarios.

In contrast to prior work, we propose a generalizable software-based framework for mmWave signal synthesis. Our \system approach comprehensively considers the physical impact of human and environmental reflections, incorporating practical factors such as the material of reflection surfaces, antenna gain properties, and multipath propagations. This makes the synthesized signal more realistic and practical. Additionally, our signal generation process does not require prior collection of mmWave signals, and the generated signals are not constrained to specific human motions or environments.

\section{Conclusion} \label{sec:cons}

This paper introduces \system, a novel and generalized software-based pipeline for mmWave signal generation. We construct physical signal transmission models to simulate human-reflected, environment-reflected, and multipath signals, enabling comprehensive full-scene signal synthesis. 
The entire generation process is software-based and requires only a snapshot of the environment for scene profiling. 
Our solution is applicable to various scenarios and downstream tasks with minimal efforts. 
%

\section*{Acknowledgment}
This work was supported by National Key R\&D Program of China 2023YFB2904000, 
the NSFC Grant No.
62372365, 
62302383,
62472346,
project funded by China Postdoctoral Science Foundation No.2023M742792, and Fundamental Research Funds for the Central Universities.
Han Ding and Cui Zhao are the corresponding authors.

\newpage
\IEEEtriggeratref{17}
\bibliographystyle{IEEEtran}
\bibliography{reference}

\end{document}